\documentclass[10pt,twocolumn,letterpaper]{article}
\usepackage{cvpr}
\usepackage{graphicx}
\usepackage{amsmath}
\usepackage{amssymb}
\usepackage{booktabs}
\usepackage{array}
\usepackage{tabu}
\usepackage{enumitem}
\usepackage{colortbl}
\usepackage[pagebackref,breaklinks,colorlinks]{hyperref}
\usepackage{multirow}
\usepackage[capitalize]{cleveref}
\crefname{section}{Sec.}{Secs.}
\Crefname{section}{Section}{Sections}
\Crefname{table}{Table}{Tables}
\crefname{table}{Tab.}{Tabs.}

\definecolor{mygray}{rgb}{0.82,0.82,0.82}

\begin{document}

\title{M$^2$-3DLaneNet: Exploring Multi-Modal 3D Lane Detection}

\author{Yueru Luo\textsuperscript{\rm 1,2} \quad Xu Yan\textsuperscript{\rm 1,2} \quad Chaoda Zheng\textsuperscript{\rm 1,2} 
\quad Chao Zheng\textsuperscript{\rm 3} \quad Shuqi Mei\textsuperscript{\rm 3} \\
\quad Tang Kun\textsuperscript{\rm 3}  \quad Shuguang Cui\textsuperscript{\rm 2,1} \quad Zhen Li\textsuperscript{\rm 2,1}\thanks{Corresponding author} \\
\textsuperscript{\rm 1} FNii, CUHK-Shenzhen \quad
\textsuperscript{\rm 2} SSE, CUHK-Shenzhen \quad
\textsuperscript{\rm 3} Tencent Map, T Lab\\
}

\newcommand{\modelname}{M$^2$-3DLaneNet}

\maketitle
\begin{abstract}
   Estimating accurate lane lines in 3D space remains challenging due to their sparse and slim nature. 
   Previous works mainly focused on using images for 3D lane detection, leading to inherent projection error and loss of geometry information.
   To address these issues, we explore the potential of leveraging LiDAR for 3D lane detection, either as a standalone method or in combination with existing monocular approaches.
   In this paper, we propose~\modelname~to integrate complementary information from multiple sensors.
   Specifically,~\modelname~lifts 2D features into 3D space by incorporating geometry information from LiDAR data through depth completion.
   Subsequently, the lifted 2D features are further enhanced with LiDAR features through cross-modality BEV fusion. 
   Extensive experiments on the large-scale OpenLane dataset demonstrate the effectiveness of~\modelname, regardless of the range (75m or 100m).
   
\end{abstract}

\section{Introduction}
\label{sec:intro}

Accurate and robust lane detection is the foundation for safety in autonomous driving.
Over the past few years, camera-based 2D lane detection has been extensively studied and achieved impressive results~\cite{gopalan2012learning, li2016deep, lee2017vpgnet, neven2018towards, zhang2018geometric, hou2019learning, liu2021condlanenet, tabelini2021keep, han2022laneformer, zheng2022clrnet}.
However, given 2D lane detection results, accurately localizing lanes in 3D space requires complex post-processing due to depth ambiguity.
\begin{figure}[t]
    \centering
    \includegraphics[width=0.9\linewidth]{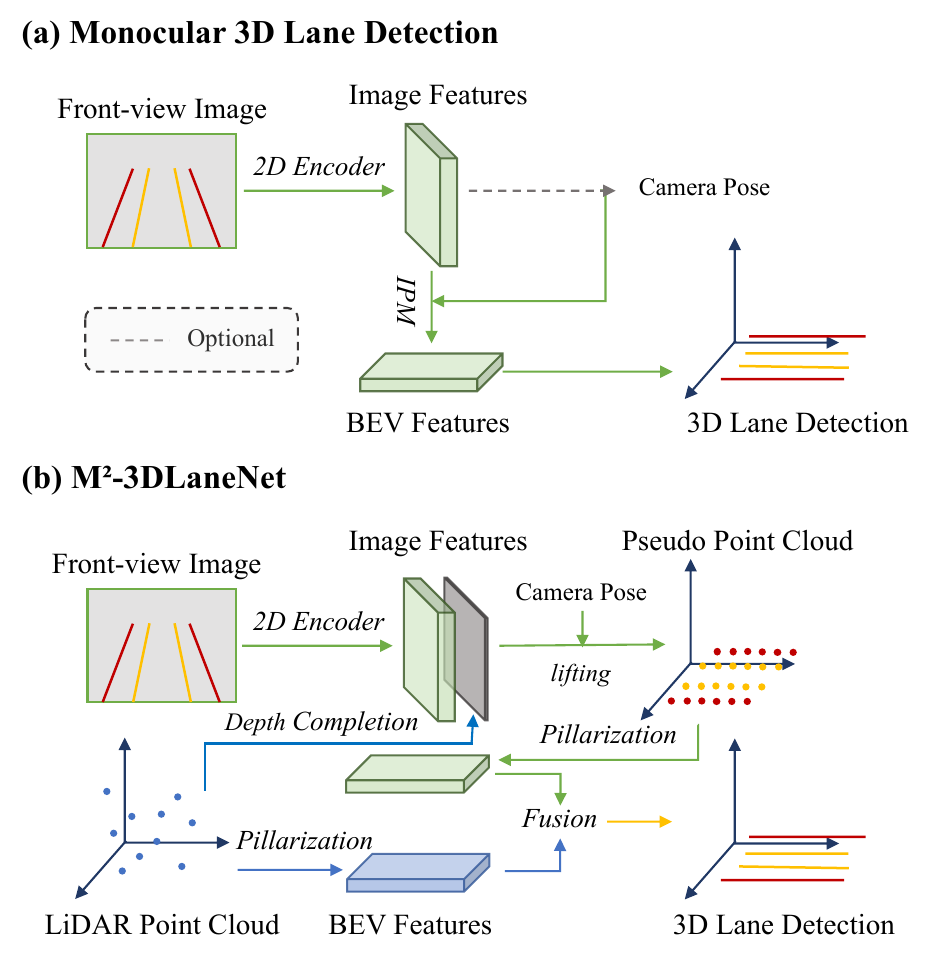}
    \vspace{-2mm}
    \caption{(a) Previous methods~\cite{garnett20193d, chen2022persformer} mainly utilize inverse perspective mapping (IPM) to transform front-view image features to BEV features, through either ground truth or learned camera poses.
    (b) Our proposed~\modelname~first lifts image features to 3D space through depth completion of LiDAR to form the pseudo point cloud, and then fuses multi-modality features in the BEV for 3D lane detection.
    }
    \label{fig:motivation}
\end{figure}
To circumvent this problem, several 3D lane open datasets~\cite{garnett20193d, guo2020gen, yan2022once} have been proposed, which have encouraged the development of algorithms~\cite{garnett20193d,efrat20203d, guo2020gen, liu2022learning, chen2022persformer, yan2022once} for the detection of 3D lanes.
Most previous methods model the 3D lane detecting in a vision-centric manner, which takes front-view camera images as inputs and predicts 3D lanes in the bird's eye view (BEV) space.
Due to the lack of depth information, these approaches mainly utilize inverse perspective mapping (IPM)~\cite{garnett20193d, efrat20203d, guo2020gen, liu2022learning} to transform image features from the front view to the BEV space, as shown in~\Cref{fig:motivation}~(a). However, IPM is based on the {flat road assumption}, which does not hold for most real-world scenarios (\eg, in the sloping terrain). To mitigate the distortion problem caused by the flat-road assumption, a recent work~\cite{chen2022persformer} adopted a deformable attention mechanism~\cite{zhu2020deformable} to optimize the view transformation problem and achieve the state-of-the-art results on 3D lane detection. Although images can provide clues about the lanes with rich semantic context,
cameras are sensitive to illumination change. Therefore, these vision-centric methods become unreliable under extreme light conditions.
Recently, LiDARs are becoming increasingly popular in autonomous driving systems. The accurate depth information provided by LiDARs, regardless of lighting variations, motivates us to explore \textit{whether LiDAR data can facilitate modern 3D lane detection and how much gain it can boost}.
Thanks to inherent rich 3D geometric information, detecting flat ground from other objects (\eg, cars, pedestrians) could be much easier in point clouds. This characteristic is preferable for lane detection, as lanes are typically located on the ground. We also find that lane lines have discriminative \emph{intensity} compared to road surfaces and other objects in most scenarios, as shown in~\Cref{fig:raw_lidar}.
However, the sparsity and slender appearance of lanes in point clouds result in lanes constituting only a small fraction of the total points. 
Consequently, lane lines in point clouds often get obscured by the surrounding ground surfaces.

Motivated by the above observation, our objective is to enhance 3D lane detection by synergistically leveraging complementary information from images and LiDAR point clouds. This paper begins with an exploration of LiDAR's capabilities in 3D lane detection. Subsequently, we introduce~\modelname, a multi-modal framework designed to achieve precise 3D lane detection, surpassing the limitations of prior single-modal approaches~\Cref{fig:motivation}~(b).~\modelname~takes geometric-rich point clouds and texture-rich images as inputs, processing them through two modal-specific streams to predict 3D lanes via multi-scale cross-modal feature fusion. 

To obtain 3D lane prediction, we propose to conduct feature fusion in a unified BEV space. In the LiDAR stream, we generate multi-scale features in BEV space using a 3D pillar-based backbone~\cite{lang2019pointpillars}. For the camera stream, the input image is initially encoded into multi-scale features and subsequently enhanced using our line-restricted feature aggregation. Afterward, the multi-scale image features are projected into 3D space as pseudo points, aided by the completed depth information obtained from LiDAR. These lifted points are pillarized into the same BEV space as the LiDAR stream. With multi-scale BEV features from both modalities in hand, we perform a bottom-up fusion, transitioning from large-scale maps to the smallest one, to generate the final 3D lane prediction.
Experiments on OpenLane dataset~\cite{chen2022persformer} show that our model surpasses all previous methods by a significant margin. 

In summary, our main contributions are twofold: 

\begin{itemize}[leftmargin=*]
\item We explored the possibility of utilizing LiDARs to detect 3D lanes among modern vision-centric solutions by conducting comprehensive experiments on Waymo\cite{waymo}-based 3D lane detection dataset, namely OpenLane. In doing so, we provide insights into how LiDARs can facilitate 3D lane detection and why their utilization is beneficial, investigating their geometric patterns and intensity—a topic that remains underexplored in the current literature.

\item We propose~\modelname, a multi-modal framework for accurate 3D lane detection, which serves as a strong baseline for future research. This framework effectively employs complementary features from camera images and LiDAR point clouds to detect 3D lanes.

\end{itemize}

\section{Related work}
\subsection{2D Lane Detection}
The goal of 2D lane detection is to detect the positions of lanes in the 2D images. Among recent methods, there are four mainstream strategies adopted to perform this task: 
1) Anchor-based methods~\cite{li2019line, torres2020keep, qin2020ultra, zheng2022clrnet} leverage line-like anchors to detect lanes out of the special nature of the slender lane.~\cite{torres2020keep} utilizes attention between anchors to aggregate global information.
Further,~\cite{qin2020ultra} uses row anchors to represent lanes, by which lane detection is formulated as a row selection problem.
2) Segmentation-based methods~\cite{hou2019learning, neven2018towards, liu2021condlanenet, wu2021yolop} aim to predict the lane mask through pixel-wise classification task. 
3) Parametric-based methods~\cite{liu2021end, feng2022rethinking, tabelini2021polylanenet} are in a characteristic way to resolve the traffic line detection, \ie, learn to predict parameters used to fit the polynomial curves of lanes.~\cite{liu2021end, tabelini2021polylanenet} model lane shape with polynomial parameters, while~\cite{feng2022rethinking} adopts Bézier curve to capture holistic lane structure.
4) Key-point-based methods~\cite{qu2021focus, wang2022keypoint} formulate lane detection problem from key points perspective and finally get lanes through associating points in the same lane.
However, since lane annotations are defined on 2D images, these methods lack the ability to accurately localize lanes in 3D space.

\begin{figure*}[t]
    \centering
    \includegraphics[width=0.9\linewidth]{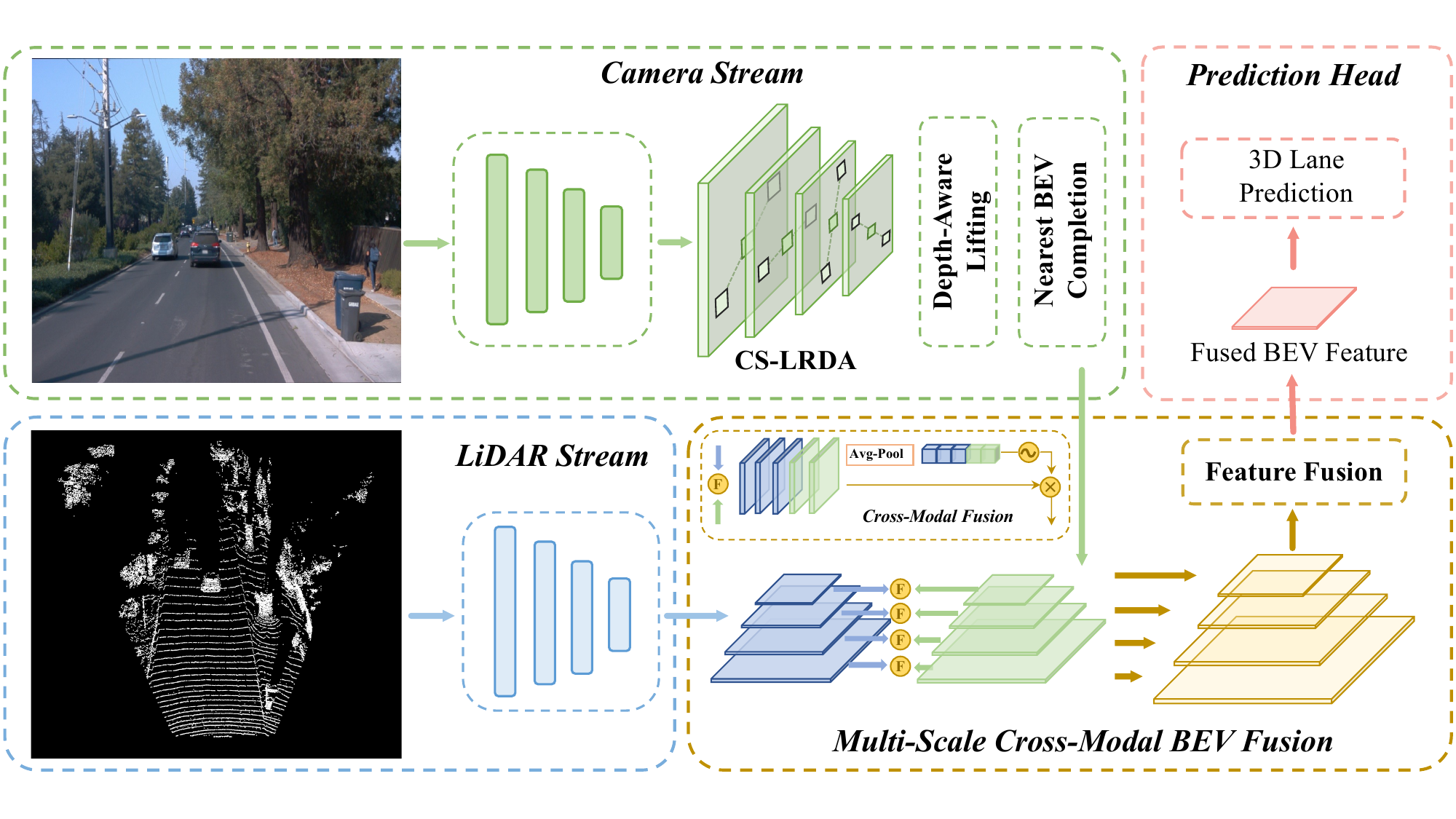}
    \vspace{-5mm}
    \caption{\textbf{Overview of the proposed~\modelname.} 
    It parallelly generates BEV features by taking an image and LiDAR point cloud as inputs, where features from the former are gained by top-down BEV generation. 
    Afterward, two BEV features are fused together with bottom-up BEV fusion, where the fused BEV features are finally used to predict 3D lanes.
    }
    \label{fig:overview}
\end{figure*}

\subsection{3D Lane Detection}
3D lane detection aims at predicting lane lines in the 3D space. 
While most deep-learning-based 3D lane detection algorithms generate 3D predictions, they still rely on monocular images due to the lack of public 3D datasets. Among these methods, 3D-LaneNet~\cite{garnett20193d} is the first to detect 3D lanes using monocular front-view images. It utilizes inverse perspective mapping (IPM) to transform the front-view images into top-view, employing camera poses predicted by a learning branch to predict lanes on a 3D plane. Gen-LaneNet~\cite{guo2020gen} proposes a virtual top view as a surrogate to address the misalignment between anchor representation and internal features projected by IPM. CLGo~\cite{liu2022learning} realizes a two-stage framework with pose estimation and polynomials estimation. These methods rely on IPM to project image features to top-view features. Recent Persformer~\cite{chen2022persformer} leverages the deformable attention mechanism to mitigate the discrepancies introduced by IPM. However, IPM introduces distortion in uphill or downhill scenarios, which jeopardizes the ability of the model to perceive the scene accurately and consistently. Alternatively, ONCE~\cite{yan2022once} directly generates 3D lanes without relying on IPM. Instead, it detects 2D lanes on images and projects them into 3D space with the help of depth estimation.
Nevertheless, these monocular approaches heavily depend on camera features, which suffer from depth ambiguity and sensitivity to light conditions. Although some approaches attempt to detect 3D lanes using LiDAR data~\cite{jung2018real, Thuy2010LANEDA, kammel2008lidar}. They often rely strongly on the hand-craft intensity threshold, making it challenging to determine suitable values across different environments.
Fortunately,~\cite{chen2022persformer} provides a large-scale 3D lane detection dataset, OpenLane, which is the first public 3D lane dataset containing both camera and LiDAR data with pixel-to-point correspondences.
This new dataset offers an opportunity to develop multi-modal approaches that could achieve better 3D lane detection.

\subsection{Bird-Eye-View Perception}
BEV-based representation draws a lot of attention among 3D perception tasks recently, owing to its strong capability and compactness in representing not only multi-sensor features but also unifying multiple tasks in a shared space. LSS~\cite{philion2020lift} proposed semantic segmentation on BEV for variable cameras by lifting 2D images to frustums. This 2D-to-3D lifting operation is accomplished with the aid of estimated depth distribution for each pixel. In this way, LSS generates a unified voxel representation from different views. Inspired by LSS~\cite{philion2020lift}, several studies~\cite{huang2021bevdet, liang2022bevfusion, liu2022bevfusion, li2022bevformer} on 3D detection have explored the capabilities of BEV representation to achieve improved performance. Among them, BEVFusion~\cite{liang2022bevfusion} fuses multi-view image features and LiDAR features into BEV space, explicitly predicting the depth distributions of image feature pixels. Similarly, another concurrent work, BEVFusion\footnote{There are two concurrent BEVFusion works in the literature.}~\cite{liu2022bevfusion} unifies 3D detection and segmentation in one framework. Different from LSS, BEVFormer~\cite{li2022bevformer} projects BEV grids into images and uses deformable attention to aggregate image features. Then, they adopt a query-based paradigm in BEV space for the final prediction.

In line with the spirit of 2D-to-3D projection, but differing in method, we lift multi-scale 2D features into LiDAR space, considering that features from monocular images are suboptimal for 3D perception. Besides, generating a large point cloud using depth distribution is computationally expensive. Thus, we employ paired LiDAR data to generate a depth map on the CPU using the image processing algorithm proposed in~\cite{ku2018defense}.

\subsection{Multi-Sensor Approaches}
As cameras and LiDARs capture complementary information, multi-sensor approaches are widely adopted in different fields. In 3D object detection, PointPainting~\cite{vora2020pointpainting} provides point clouds with their corresponding 2D semantics. Taking advantage of the attention mechanism,~\cite{li2022deepfusion, chen2022autoalign, bai2022transfusion} adaptively model the 2D-3D mapping through multi-modal fusion. SFD~\cite{wu2022sparse} proposes an RoI fusion strategy to aggregate multi-modal RoI features and designs color point convolution to extract pseudo point cloud features. Regarding 3D lane detection, there are few previous approaches related to multi-sensor approaches due to the lack of a dataset. Early work~\cite{bai2018deep} adopts multi-modal input on their private dataset, but it relies on supervision from additional high-definition maps. Furthermore, they did not explore geometric information in LiDAR point clouds and solely extract features through 2D-CNNs using a three-channel image obtained from raw LiDAR points.

\section{Method}
In this section, we present the architecture of~\modelname, designed to effectively leverage multi-modal features for accurate 3D lane detection. We begin by describing the overall structure of our~\modelname~\Cref{sec:arch}. Then, we explain the process of lifting 2D image features into 3D space in~\Cref{sec:top-down-bev}. Following, we elaborate on the fusion of camera and LiDAR features into the unified 3D space in~\Cref{sec:mmfusion}. By aggregating the two-modal information in a shared BEV space, we achieve a comprehensive representation for the final prediction.

\subsection{Two-Stream Architecture} \label{sec:arch}
As shown in Figure~\ref{fig:overview}, our architecture comprises two parallel streams to generate multi-scale features for each modality.

\noindent\textbf{Camera Stream.} 
The camera stream processes RGB images and encodes them into multi-scale BEV feature maps supported by our proposed depth-aware lifting. Following~\cite{chen2022persformer}, we adopt the first meta-stage of EfficientNet-B7~\cite{tan2019efficientnet} followed by eight convolution layers as our image backbone, producing four different scales of features. These multi-scale features are lifted into 3D space, leveraging their corresponding real depth maps from LiDAR, and finally encoded into BEV space.

\noindent\textbf{LiDAR Stream.} 
In the LiDAR stream, we encode LiDAR point cloud using PointPillars~\cite{lang2019pointpillars}.
This process involves dividing the point cloud into different pillars and encoding each pillar into a high-dimensional feature vector using a mini-PointNet~\cite{qi2017pointnet}. Subsequently, all pillars are mapped to the 2D BEV grid based on their corresponding locations. Then, we use CNNs to generate LiDAR BEV features in four scales, similar to the camera stream.

\subsection{Camera Stream: Front View to BEV}\label{sec:top-down-bev}
\begin{figure}[t]
    \centering
    \includegraphics[width=0.8\linewidth]{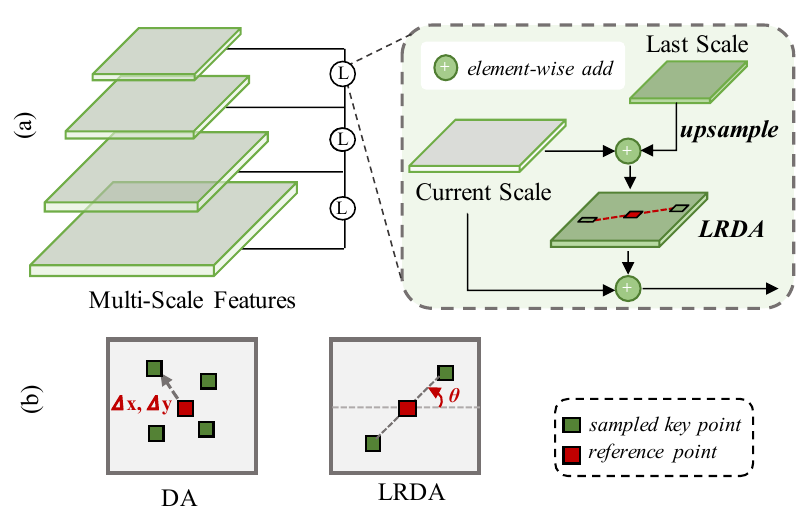}
    \vspace{-3mm}
    \caption{\textbf{Cross-Scale Line-restricted Deformable Attention (CS-LDA).} 
    (a) The workflow of CS-LDA, in which the features from two different scales are merged for predicting the current sampling position, and a residual connection are added after attention enhancement.
    (b) The comparison of previous deformable attention (DA) with our line-restricted deformable attention (LDA). The main difference lies in that we restrict the sampling locations with an explicit slope–intercept line form.
    Since this operation is applied from the top to the bottom, the input of the first layer is just itself.
    }
    \label{fig:CS-LDA}
\end{figure}

To obtain multi-scale BEV features from the input multi-scale features of the image, we enhance the 2D feature maps using our newly proposed Cross-Scale Line-restricted Deformable Attention (CS-LDA) module, as depicted in~\Cref{fig:CS-LDA}. CS-LDA is a modification of the deformable attention~\cite{zhu2020deformable} that restricts sampling locations along straight lines, considering the long and slim nature of lane lines. 
Additionally, it propagates high-level features to low-level ones in a coarse-to-fine paradigm to predict sampling locations.

Our experiments reveal that the line restriction in CS-LDA leads to a 0.2 improvement in the final F-Score without significantly increasing complexity, as shown in Table~\ref{tab:ablation_components}.

After enhancing the front-view features, we transform them into BEV space using the following steps:
1) \emph{Lifting} to 3D space and creating pseudo point clouds (see~\Cref{subsubsec:lifting}).
2) \emph{Pillarizing} the lifted pseudo point clouds and generating BEV representations, similar to the process used in the LiDAR stream.
3) \emph{Completing} the BEV grid to alleviate the effects of occlusion and limited field-of-view (see~\Cref{subsec:nbc}).

\subsubsection{Depth-aware Lifting} \label{subsubsec:lifting}

\begin{figure}[t]
    \centering
    \includegraphics[width=0.9\linewidth]{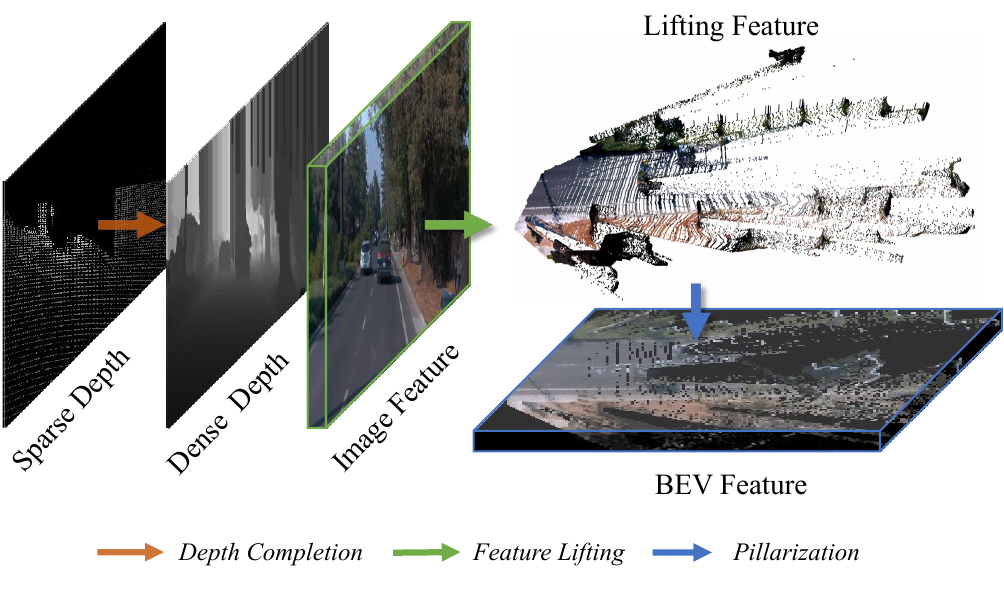}
    \vspace{-4mm}
    \caption{\textbf{Depth-aware Lifting.} 
    To provide a clearer visualization of the lifting process, we use the thick image to represent its corresponding features. The lifting process involves the following steps: 1) Completing the sparse depth map to obtain a dense depth map.
    2) Lifting multi-scale image features into 3D space based on the dense depth map.
    3) Conducting pillarization to obtain the BEV features from the 3D space.
    }
    \vspace{-4mm}
    \label{fig:bev}
\end{figure}

To lift the features to 3D space, we first need the corresponding depth of each feature pixel. Given a LiDAR point cloud $\mathbf{P} \in \mathbb{R}^{N \times 3}$, we first project it onto the image plane, obtaining a sparse depth map $\mathbf{D}$. This occurs because the original LiDAR points only map to a small proportion of pixels in the image. Then, we apply an efficient depth completion algorithm~\cite{ku2018defense}, specifically designed for completing sparse LiDAR-based depth maps using image processing techniques. This algorithm generates a dense depth map $\mathbf{\hat{D}}$. With the dense depth map in hand, we can then transform the features from the image plane into 3D space, as shown in~\Cref{fig:bev}.

In practice, given the input image with the size of $(H, W)$, we first generate image coordinates $\mathbb{C}$ according to the dense depth map $\mathbf{\hat{D}}$:
\begin{equation}
    \begin{aligned}
    \mathbb{C} = \{(u, v, d) |~u \in [1, W], v \in [1, H]\},
    \end{aligned}
\end{equation}
where $d = \mathbf{\hat{D}}_{uv}$.
Afterwards, we transform the image coordinates $\mathbb{C}$ into the 3D space, utilizing the camera intrinsic and extrinsic matrices $K \in \mathbb{R}^{4 \times 4}$ and $T \in \mathbb{R}^{4 \times 4}$. Specifically, given the $i$-th image coordinate $\mathbb{C}_i = (u_i, v_i, d_i)$, its corresponding coordinate $(x_i, y_i, z_i)$ in the world system can be calculated as:
\begin{small}
\begin{equation}
\begin{aligned}
\relax
    [x_i, y_i, z_i, 1]^\mathsf{T} = T^{-1} \cdot K^{-1} \cdot [u_i \times d_i, v_i \times d_i, d_i, 1]^\mathsf{T}.
\end{aligned}
\end{equation}
\end{small}
After the above transformation, we lift the multi-scale image features into the 3D space. To unify the image features into a common space, we perform pillarization~\cite{lang2019pointpillars} on the above multi-scale features, obtaining multi-scale BEV features, as shown in~\Cref{fig:bev}. 

\subsubsection{Nearest BEV Completion.} \label{subsec:nbc}
Due to occlusion and limited field-of-view, the BEV maps derived from the camera stream may contain noticeable empty grids. To tackle this problem, we apply the Nearest BEV Completion to fill the empty BEV areas. For each BEV feature map, we first generate an occupancy map, which records the occupancy status of the current lattice cells. After that, for each empty lattice, we perform interpolation using its nearest neighbor. Meanwhile, we record the offset and Euclidean distance to its nearest neighbor and use them as additional features concatenated with the original BEV features.

\subsection{Multi-Scale Cross-Modal BEV Fusion} \label{sec:mmfusion}
We employ a bottom-up fusion approach to integrate the multi-modality features from the camera and LiDAR streams, as demonstrated in~\Cref{fig:overview}. For each scale, we begin by concatenating the two feature maps obtained from different modalities. Subsequently, we apply the channel-wise attention~\cite{hu2018squeeze} for single-scale fusion, enhancing the combined feature representation. Afterward, we aggregate multi-scale cross-modal features from bottom to up, culminating in a single fused BEV feature map. This integrated feature map serves as the foundation for our final predictions.

\subsection{Prediction and Objective}
We adopt an anchor-based approach, following~\cite{chen2022persformer}, to detect 3D lanes.
Given an image and its ground-truth annotations $\mathbf{Y}^{gt}_{(\cdot)}$,
the overall objective function between prediction and ground truth is formulated as follows:
\begin{equation}
    \begin{aligned}
    \mathcal{L}_{all} =& \lambda_{1}\mathcal{L}_{3d}(\mathbf{Y}^{\prime}_{3d}, \mathbf{Y}_{3d}^{gt}) + \lambda_{2}\mathcal{L}_{2d}(\mathbf{Y}^{\prime}_{2d}, \mathbf{Y}_{2d}^{gt}) \ +\\
    &\lambda_{3}\mathcal{L}_{seg}(\mathbf{Y}_{b}^{\prime}, \mathbf{Y}_{b}^{gt}),
\end{aligned}
\end{equation}
where the 3D prediction $\mathbf{Y}^{\prime}_{3d}$ contains three components: 
1) regression $\mathbf{Y}_{3d}^{\prime_{reg}} \in \mathbb{R}^{ N_{anchor} \times N_s\times 2}$;
2) category, $\mathbf{Y}^{\prime_{cls}}_{3d} \in \mathbb{R}^{N_{anchor} \times 1}$; 
3) visibility $\mathbf{Y}^{\prime_{vis}}_{3d} \in \mathbb{R}^{ N_{anchor} \times N_s \times 1}$. $N_s$ denotes the number of sampled points along the $y$-axis, which is predefined and remains the same for each anchor. $\mathbf{Y}_{3d}^{\prime_{reg}}$ denotes the predicted offsets to predefined anchors, trained by optimizing a Smooth-L1 loss. $\mathbf{Y}_{cls}^{3d}$ and $\mathbf{Y}_{vis}^{3d}$ denote the category and visibility of anchors respectively. Notably, in the OpenLane dataset, \textit{visibility} for each point is used solely to indicate the validity of the point on each lane, due to its label generation processing, \eg, the lanes in the sky are filtered. To enhance the capability of 2D features, we adopt auxiliary losses of 2D lane detection, which consist of similar components as the 3D lane detection, \ie, {regression, category, and visibility}. Moreover, a binary semantic segmentation loss is employed on the fused BEV feature supervised by the projection of 3D lane annotation.

\begin{table*}[t]
\small
    \centering
    \caption{Comparison with other 3D lane detection methods on the OpenLane \textit{different scenarios} with \textbf{1.5m} and \textbf{0.5m} \textit{Dist.}(distance threshold) under \textbf{100m} and \textbf{75m} range evaluation settings,
    Note that the maximum reach of the LiDAR system in Waymo~\cite{waymo} is 76.8 meters.
    L and C represent LiDAR and Camera respectively.
    \textbf{M$^2$ C-Stream$^\star$:} our model without the LiDAR-Stream. Depth maps in the camera stream are acquired based on LiDAR data, making this model a C$+$L modality approach.
    \textbf{Persformer $+$ LiDAR$^\dagger$:} We equip Persformer~\cite{chen2022persformer} with our LiDAR branch and multi-scale fusion module.
    \textbf{Persformer$^\ast$:} Persformer~\cite{chen2022persformer} evaluated on the 75m range setting using their provided checkpoint.
    The improvements are compared against the previous state-of-the-art Persformer~\cite{chen2022persformer} model.
    }
    \vspace{-2mm}
    \begin{tabu}{c!{\vrule width 1.2pt}c!{\vrule width 1.2pt}c|c!{\vrule width 1.2pt}c!{\vrule width 1.2pt}c|c|c|c|c|c}
    \toprule[1.2pt]
    \multirow{2}{*}{\textit{Dist.}} & \multirow{2}{*}{Range} & \multirow{2}{*}{Methods} & \multirow{2}{*}{Modality} & \multirow{2}{*}{All} & Up \&  & \multirow{2}{*}{Curve} & Extreme & \multirow{2}{*}{Night} & \multirow{2}{*}{Intersection} & Merge \& \\
    & & & & &  Down & & Weather & & & Split
    \\
    \hline \hline
    \multirow{12}{*}{{\rotatebox{90}{\small{\textbf{1.5 m}}}}} &
    \multirow{7}{*}{{\rotatebox{90}{\small{\textbf{100 m}}}}} 
    & 3DLaneNet~\cite{garnett20193d} & C & 44.1 & 40.8 & 46.5 & 47.5 & 41.5 & 32.1 & 41.7 \\
    & & GenLaneNet~\cite{guo2020gen} & C & 32.3 & 25.4 & 33.5 & 28.1 & 18.7 & 21.4 & 31.0 \\
    & & Persformer~\cite{chen2022persformer} & C & \underline{50.5} & \underline{42.4} & \underline{55.6} & \underline{48.6} & \underline{46.6} & \underline{40.0} & \underline{50.7} \\
    & & M$^2$ C-Stream$^\star$ & C+L & 50.9 & 50.4 & 60.0 & 49.0 & 50.3 & 43.5 & 48.2 \\
    & & Persformer + LiDAR$^\dagger$ & C+L & 53.0 & 45.5 & 58.8 & 53.8 & 49.0 & 52.6 & 40.9 \\
    & & M$^2$-3DLaneNet (ours) & C+L & \textbf{55.5} & \textbf{53.4} & \textbf{60.7} & \textbf{56.2} & \textbf{51.6} & \textbf{43.8} & \textbf{51.4} \\
    & & \emph{Improvement} & - 
    & \textcolor[rgb]{0.0,0.5,0.8}{\textit{$\uparrow$5.0}}
    & \textcolor[rgb]{0.0,0.5,0.8}{\textit{$\uparrow$11.0}}
    & \textcolor[rgb]{0.0,0.5,0.8}{\textit{$\uparrow$5.1}}
    & \textcolor[rgb]{0.0,0.5,0.8}{\textit{$\uparrow$7.6}}
    & \textcolor[rgb]{0.0,0.5,0.8}{\textit{$\uparrow$5.0}}
    & \textcolor[rgb]{0.0,0.5,0.8}{\textit{$\uparrow$3.8}}
    & \textcolor[rgb]{0.0,0.5,0.8}{\textit{$\uparrow$0.7}}
    \\
    \cline{2-11}
    \addlinespace[1.2pt]
    &
    \multirow{5}{*}{{\rotatebox{90}{\small{\textbf{75 m}}}}}
    & Persformer~\cite{chen2022persformer}$^\ast$ & C & \underline{53.3} & \underline{43.6} & \underline{56.7} & \underline{56.9} & \underline{48.9} & \underline{40.2} & \underline{54.0} \\
    & & M$^2$ C-Stream$^\star$ & C+L & 59.2 & 52.3 & 63.8 & 60.2 & 57.8 & 47.9 & 56.9 \\
    & & Persformer + LiDAR$^\dagger$ & C+L & 56.9 & 47.6 & 61.3 & 58.0 & 56.6 & 45.4 & 55.4 \\
    & & M$^2$-3DLaneNet (ours) & C+L & \textbf{60.5} & \textbf{53.8} & \textbf{65.6} & \textbf{61.2} & \textbf{59.0} & \textbf{49.5} & \textbf{57.8} \\
    & & \emph{Improvement} & - 
    & \textcolor[rgb]{0.0,0.5,0.8}{\textit{$\uparrow$7.2}}
    & \textcolor[rgb]{0.0,0.5,0.8}{\textit{$\uparrow$10.2}}
    & \textcolor[rgb]{0.0,0.5,0.8}{\textit{$\uparrow$8.9}}
    & \textcolor[rgb]{0.0,0.5,0.8}{\textit{$\uparrow$4.3}}
    & \textcolor[rgb]{0.0,0.5,0.8}{\textit{$\uparrow$10.1}}
    & \textcolor[rgb]{0.0,0.5,0.8}{\textit{$\uparrow$9.3}}
    & \textcolor[rgb]{0.0,0.5,0.8}{\textit{$\uparrow$3.8}}
    \\
    \tabucline[1.2pt]{-}
    \addlinespace[1.2pt]
    \multirow{8}{*}{{\rotatebox{90}{\small{\textbf{0.5 m}}}}} &
    \multirow{4}{*}{{\rotatebox{90}{\small{\textbf{100 m}}}}} 
    & Persformer~\cite{chen2022persformer}$^\ast$ & C & \underline{32.0} & \underline{28.0} & \underline{32.5} & \underline{35.7} & \underline{28.9} & \underline{23.6} & \underline{30.4} \\
    & & Persformer + LiDAR$^\dagger$ & C+L & 42.6 & 36.2 & 42.4 & 44.1 & 39.7 & 31.9 & 42.1 \\
    & & M$^2$-3DLaneNet (ours) & C+L & \textbf{48.2} & \textbf{37.8} & \textbf{45.5} & \textbf{46.5} & \textbf{43.0} & \textbf{36.0} & \textbf{42.1} \\
    & & \emph{Improvement} & - 
    & \textcolor[rgb]{0.0,0.5,0.8}{\textit{$\uparrow$16.2}}
    & \textcolor[rgb]{0.0,0.5,0.8}{\textit{$\uparrow$9.8}}
    & \textcolor[rgb]{0.0,0.5,0.8}{\textit{$\uparrow$13.0}}
    & \textcolor[rgb]{0.0,0.5,0.8}{\textit{$\uparrow$10.8}}
    & \textcolor[rgb]{0.0,0.5,0.8}{\textit{$\uparrow$14.1}}
    & \textcolor[rgb]{0.0,0.5,0.8}{\textit{$\uparrow$12.4}}
    & \textcolor[rgb]{0.0,0.5,0.8}{\textit{$\uparrow$11.7}}
    \\
    \cline{2-11}
    \addlinespace[1.2pt]
    & \multirow{4}{*}{{\rotatebox{90}{\small{\textbf{75 m}}}}}
    & Persformer~\cite{chen2022persformer}$^\ast$ & C & \underline{35.3} & \underline{29.6} & \underline{35.1} & \underline{40.2} & \underline{32.4} & \underline{26.3} & \underline{33.4} \\
    & & Persformer + LiDAR$^\dagger$ & C+L & 45.1 & 37.5 & 44.3 & 49.3 & 45.3 & 35.2 & 42.8 \\
    & & M$^2$-3DLaneNet (ours) & C+L & \textbf{53.8} & \textbf{46.7} & \textbf{55.4} & \textbf{54.1} & \textbf{53.4} & \textbf{42.6} & \textbf{50.9} \\
    & & \emph{Improvement} & - 
    & \textcolor[rgb]{0.0,0.5,0.8}{\textit{$\uparrow$18.5}}
    & \textcolor[rgb]{0.0,0.5,0.8}{\textit{$\uparrow$17.1}}
    & \textcolor[rgb]{0.0,0.5,0.8}{\textit{$\uparrow$20.3}}
    & \textcolor[rgb]{0.0,0.5,0.8}{\textit{$\uparrow$13.9}}
    & \textcolor[rgb]{0.0,0.5,0.8}{\textit{$\uparrow$21.0}}
    & \textcolor[rgb]{0.0,0.5,0.8}{\textit{$\uparrow$16.3}}
    & \textcolor[rgb]{0.0,0.5,0.8}{\textit{$\uparrow$17.5}}
    \\
    \toprule[1.2pt]
    \end{tabu}
    \label{tab:main_result}
\end{table*}

\begin{table*}[t]
\small
    \centering
        \caption{Comprehensive comparison with other 3D lane methods on OpenLane in F-score and \textit{errors} along both X and Z axes (\textit{meter}).}
    \vspace{-2mm}
    \begin{tabu}{c!{\vrule width 1.2pt}c!{\vrule width 1.2pt}c|c!{\vrule width 1.2pt}c!{\vrule width 1.2pt}c|c|c|c}
    \toprule[1.2pt]
    \textit{Dist.} & Range & Methods & Modality & F-Score$\uparrow$ & X error near$\downarrow$ & X error far$\downarrow$ & Z error near$\downarrow$ & Z error far$\downarrow$ \\
    \hline \hline
    \multirow{12}{*}{{\rotatebox{90}{\small{\textbf{1.5 m}}}}} &
    \multirow{7}{*}{{\rotatebox{90}{\small{\textbf{100 m}}}}} 
    & 3DLaneNet~\cite{garnett20193d} & C & 44.1 & 0.479 & 0.572 & 0.367 & 0.443 \\
    & & GenLaneNet~\cite{guo2020gen} & C & 32.3 & 0.591 & 0.684 & 0.411 & 0.521 \\
    & & Persformer~\cite{chen2022persformer} & C & \underline{50.5} & \underline{0.485} & \underline{0.553} & \underline{0.364} & \underline{0.431} \\
    & & M$^2$ C-Stream$^\star$ & C+L & 50.9 & 0.433 & 0.536 & 0.333 & 0.452  \\
    & & Persformer + LiDAR$^\dagger$ & C+L & 53.0 & 0.458 & 0.519 & 0.357 & 0.416 \\
    
    & & M$^2$-3DLaneNet (ours) & C+L & \textbf{55.5} & \textbf{0.431} & \textbf{0.487} & \textbf{0.327} & \textbf{0.401}   \\
    & & \emph{Improvement} & - 
    & \textcolor[rgb]{0.0,0.5,0.8}{\textit{$\uparrow$5.0}}
    & \textcolor[rgb]{0.0,0.5,0.8}{\textit{$\downarrow$0.054}}
    & \textcolor[rgb]{0.0,0.5,0.8}{\textit{$\downarrow$0.066}}
    & \textcolor[rgb]{0.0,0.5,0.8}{\textit{$\downarrow$0.037}}
    & \textcolor[rgb]{0.0,0.5,0.8}{\textit{$\uparrow$0.030}}
    \\
    \cline{2-9}
    \addlinespace[1.2pt]
    & \multirow{5}{*}{{\rotatebox{90}{\small{\textbf{75 m}}}}} 
    & Persformer$^\ast$ & C & \underline{53.3} & \underline{0.478} & \underline{0.563} & \underline{0.363} & \underline{0.408} \\
    & & M$^2$ C-Stream$^\star$ & C+L & 59.2 & 0.402 & 0.420 & 0.313 & 0.335 \\
    & & Persformer + LiDAR$^\dagger$ & C+L & 56.9 & 0.452 & 0.467 & 0.354 & 0.357 \\
    & & M$^2$-3DLaneNet (ours) & C+L & \textbf{60.5} & \textbf{0.396} & \textbf{0.405} & \textbf{0.311} & \textbf{0.326}  \\
    & & \textit{Improvement} & - 
    & \textcolor[rgb]{0.0,0.5,0.8}{\textit{$\uparrow$7.2}}
    & \textcolor[rgb]{0.0,0.5,0.8}{\textit{$\downarrow$0.082}}
    & \textcolor[rgb]{0.0,0.5,0.8}{\textit{$\downarrow$0.158}}
    & \textcolor[rgb]{0.0,0.5,0.8}{\textit{$\downarrow$0.052}}
    & \textcolor[rgb]{0.0,0.5,0.8}{\textit{$\downarrow$0.082}}
    \\
    \tabucline[1.2pt]{-}
    \addlinespace[1.2pt]
    \multirow{8}{*}{{\rotatebox{90}{\small{\textbf{0.5 m}}}}} &
    \multirow{4}{*}{{\rotatebox{90}{\small{\textbf{100 m}}}}} 
    & Persformer~\cite{chen2022persformer}$^\ast$ & C & \underline{32.0} & \underline{0.459} & \underline{0.557} & \underline{0.374} & \underline{0.464} \\
    & & Persformer + LiDAR$^\dagger$ & C+L & 42.6 & 0.431 & 0.488 & 0.355 & 0.411 \\
    & & M$^2$-3DLaneNet (ours) & C+L & \textbf{48.2} & \textbf{0.399} & \textbf{0.459} & \textbf{0.325} & \textbf{0.397}   \\
    & & \emph{Improvement} & - 
    & \textcolor[rgb]{0.0,0.5,0.8}{\textit{$\uparrow$16.2}}
    & \textcolor[rgb]{0.0,0.5,0.8}{\textit{$\downarrow$0.060}}
    & \textcolor[rgb]{0.0,0.5,0.8}{\textit{$\downarrow$0.098}}
    & \textcolor[rgb]{0.0,0.5,0.8}{\textit{$\downarrow$0.049}}
    & \textcolor[rgb]{0.0,0.5,0.8}{\textit{$\downarrow$0.067}}
    \\
    \cline{2-9}
    \addlinespace[1.2pt]
    & \multirow{4}{*}{{\rotatebox{90}{\small{\textbf{75 m}}}}} 
    & Persformer$^\ast$ & C & \underline{35.3} & \underline{0.438} & \underline{0.469} & \underline{0.361} & \underline{0.374} \\
    & & Persformer + LiDAR$^\dagger$ & C+L & 45.1 & 0.424 & 0.426 & 0.351 & 0.347 \\
    & & M$^2$-3DLaneNet (ours) & C+L & \textbf{53.8} & \textbf{0.375} & \textbf{0.381} & \textbf{0.309} & \textbf{0.322}  \\
    & & \textit{Improvement} & - 
    & \textcolor[rgb]{0.0,0.5,0.8}{\textit{$\uparrow$18.5}}
    & \textcolor[rgb]{0.0,0.5,0.8}{\textit{$\downarrow$0.063}}
    & \textcolor[rgb]{0.0,0.5,0.8}{\textit{$\downarrow$0.088}}
    & \textcolor[rgb]{0.0,0.5,0.8}{\textit{$\downarrow$0.052}}
    & \textcolor[rgb]{0.0,0.5,0.8}{\textit{$\downarrow$0.052}}
    \\
    \toprule[1.2pt]
    \end{tabu}
    \vspace{-4mm}
    \label{tab:main_result_error}
\end{table*}

\label{sec:exp}
\section{Experiment}
As OpenLane~\cite{chen2022persformer} is currently the only publicly available dataset that includes LiDAR-camera inputs with their correspondence, we conducted all experiments using this dataset.

\subsection{Implementation Details}
For 2D image feature extraction, we adopt the same structure as Persformer~\cite{chen2022persformer}, as illustrated in Section \ref{sec:arch}. For point clouds, we follow PointPillar~\cite{lang2019pointpillars} to generate BEV features for both LiDAR data and lifted image feature point clouds. All models are trained with batch size $16$ and overall epochs $56$.  For a fair comparison, we resize input images to the same resolution as previous methods, \ie, $480 \times 360$, during both the training and testing phases. The data augmentation includes only image random rotation from $-10^\circ$ to $10^\circ$. Through all experiments, we use the same hyperparameters and training protocols. Specifically, we use AdamW~\cite{loshchilov2017decoupled} as the optimizer, the learning rate is set to $2\times 10^{-4}$ at the beginning of training, and we use a cosine annealing scheduler~\cite{loshchilov2016sgdr} with $T_{max}=8$ to update the learning rate. 

\subsection{Datasets}
\noindent \textbf{OpenLane.}
Built on Waymo Open Dataset~\cite{sun2020scalability}, OpenLane contains $200$K frames and $880$K annotated lanes. Overall $1000$ segments are split into \texttt{train} and \texttt{val} sets as the original Waymo partition. Each segment of Waymo LiDAR data~\cite{sun2020scalability} is sampled at 10Hz for 20s with 64-beam LiDARs. Since OpenLane~\cite{chen2022persformer} annotates only the lanes in the front view, we exclude points in the LiDAR data that fall outside the field of the front view. Additionally, OpenLane provides category information for each lane, \eg, {white dash line}, and road curbside, resulting in a total of 14 categories. 

\subsection{Evaluation Metrics}\label{subsec:eval_metric}
Following the official metrics of OpenLane~\cite{chen2022persformer}, the evaluation of 3D lane detection is formulated as a matching problem between predictions and ground truth based on \textit{edit distance}. After matching, the corresponding metrics can be calculated in terms of F-Score, category accuracy, and X/Z error over the matched lanes. A correct prediction requires the point-wise distances of 75$\%$  points between the prediction and ground truth to be less than the maximum allowed distance, 1.5m. The perception distance is set as [3m, 103m] for Y-axis, dubbed the 100m range setting here.

We first evaluate models using the official metrics, 1.5m distance threshold, and 100m range setting. Further, to better exploit the capability of point clouds for 3D lane detection, we adopt 75m as the farthest perception distance, which falls within the valid range of LiDAR data in Waymo, dubbed 75m range setting (see~\Cref{tab:main_result}~\ref{tab:main_result_error}). LiDAR-based and multi-modal methods (see~\Cref{tab:mm_result_comp}) are mainly investigated under this setting. Moreover, we examined the performance of models with a more restrictive point-wise distance threshold, \ie, reducing it from the original 1.5m to \textbf{0.5m} (the lower part of~\Cref{tab:main_result}~\ref{tab:main_result_error}).
This is because we believe that a 1.5m bias would seriously hamper the safety of real-world autonomous driving. 

\subsection{Main Results} \label{sec:main_result}
We compare our approach with existing state-of-the-art 3D lane detection methods on OpenLane~\cite{chen2022persformer} dataset over different metrics.~\Cref{tab:main_result} shows the F-Score comparison over \texttt{val} set and six challenging scenarios.~\Cref{tab:main_result_error} illustrates more comprehensive evaluation result comparisons on \texttt{val} set. Our proposed~\modelname~surpasses its competitors by a significant margin when evaluated on a variety of scenarios and metrics.

\begin{figure}[ht]
    \centering
    \includegraphics[width=1.0\linewidth]{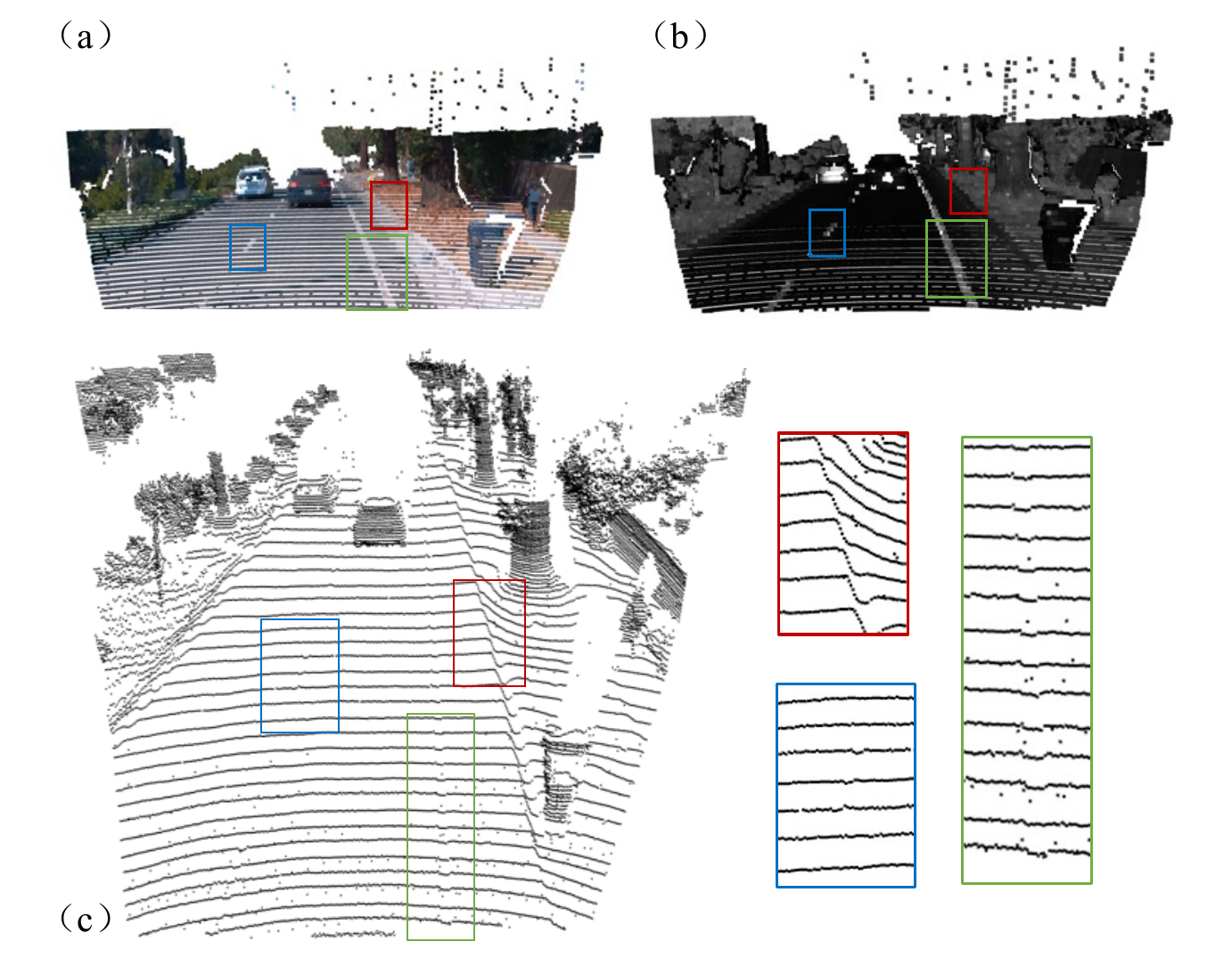}
    \vspace{-6mm}
    \caption{\textbf{Why LiDAR works?} 
    (a) shows the point cloud with RGB retrieved from the image and (b) shows the point cloud with intensity. 
    (c) demonstrate the raw point cloud, in which we zoom in on three specific areas marked with red, blue, and green boxes. It can be found that the shape of the point cloud geometry alters at spots where lane lines and road curbs are present.
    }
    \vspace{-3mm}
    \label{fig:raw_lidar}
\end{figure}
\subsubsection{Comparison with monocular methods}
\smallskip
\begin{figure*}[htbp]
    \centering
    \includegraphics[width=0.9\linewidth]{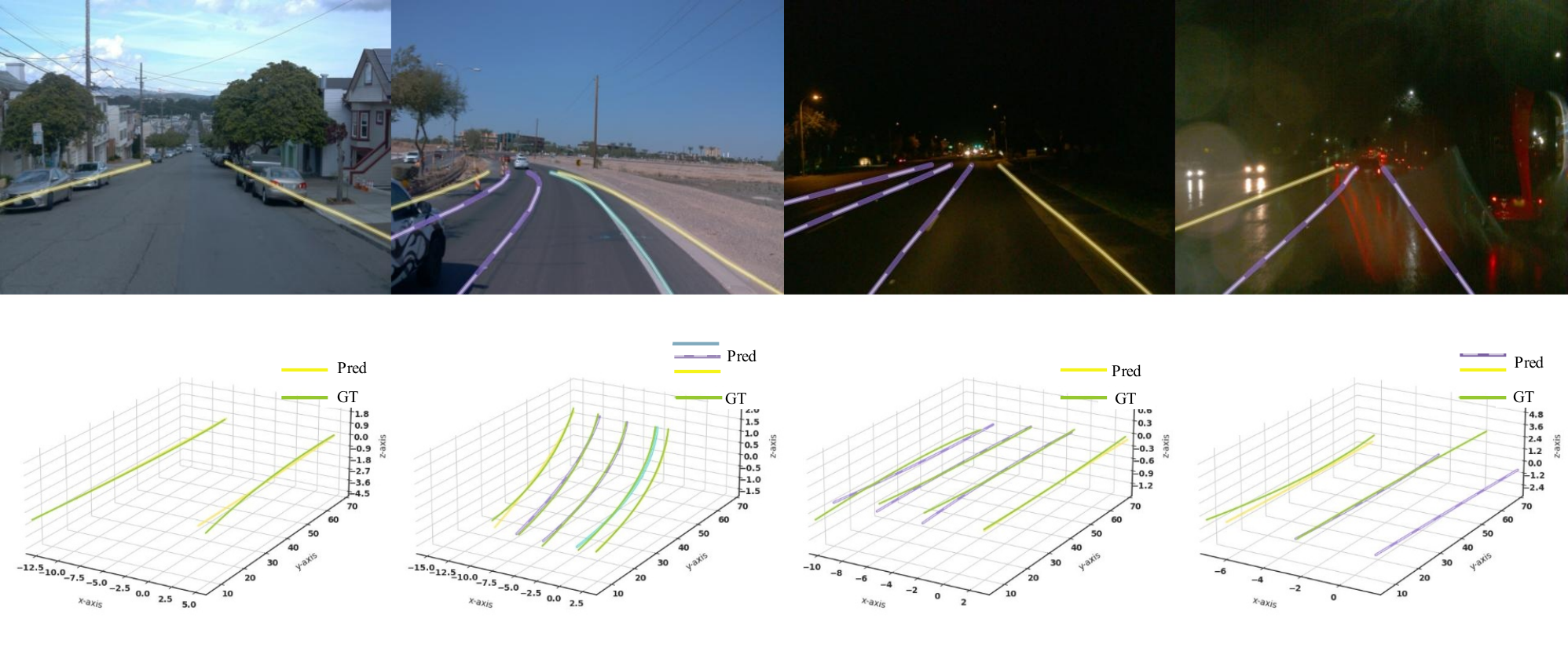}
    \vspace{-6mm}
    \caption{\textbf{Visualization.} We show the lane detection of~\modelname~in the 3D space and the corresponding image in various scenarios, where the green lines represent ground truth and the other colors are determined by different categories. Notably, the yellow line denotes the predicted road curbside, which is a class in the OpenLane dataset.}
    \vspace{-2mm}
    \label{fig:visualization}
\end{figure*}

\noindent \textbf{100m Range Evaluation.}
As shown in~\Cref{tab:main_result}, although the point clouds are missing from 75m to 103m, our method still surpasses the previous SOTA by a clear margin of 5.0$\%$ in F-Score, 0.054m/0.066m in X near/far error and 0.037m/0.030m in Z near/far error. The significant improvements demonstrate the effectiveness of our method in exploiting information from point clouds and integrating with semantics from the image for accurate 3D lane detection.

\noindent \textbf{75m Range Evaluation.}
As shown in~\Cref{tab:mm_result_comp}, our LiDAR stream (LiDAR$^\dagger$) outperforms the previous state-of-the-art method Persformer~\cite{chen2022persformer} across all metrics on OpenLane. Equipping LiDAR$^\dagger$ with our camera stream,~\modelname~achieves higher performance than previous SOTA across all metrics. Specifically, we observe a remarkable 7.2$\%$ improvement in F-Score, as well as reductions of 0.082m/0.0158m in X near/far error and reductions of 0.052m/0.082m in Z near/far error.

\noindent \textbf{More Restrict Distance Threshold.}
As discussed in~\Cref{subsec:eval_metric}, we propose adopting a smaller distance threshold that is more suitable for real-world autonomous driving scenarios. Therefore, we employ a more restrictive threshold of \textbf{0.5m} to evaluate the performance of different models. As shown in Table~\ref{tab:main_result}~\ref{tab:main_result_error}, when using a more challenging threshold, the margins between our results and the previous SOTA(Persformer~\cite{chen2022persformer}) become significantly larger across different setups. Specifically, under 100m range setting, our~\modelname~surpasses Persformer~\cite{chen2022persformer} by \textbf{16.2$\%$} in F-Score, even when we miss points beyond 75m.
Furthermore, with the valid LiDAR range (75m range setting), our~\modelname~achieves even greater improvements, showing an \textbf{18.5$\%$ } increase in F-Score on the overall \texttt{val} set and consistently outperforming the previous SOTA method by remarkable margins in all six challenging scenarios.
These strong performance results further demonstrate the strength and effectiveness of our design.

\noindent \textbf{Effect of LiDAR.}
As evident from the results in Table~\ref{tab:main_result}, our model achieves significant gains over six challenging scenes. For instance, when using a 1.5m distance threshold,~\modelname~shows improvements of 10.1$\%$ /10.2$\%$  and 5.0$\%$ /11.0$\%$  in F-Score compared to the existing SOTA under the 100m/75m setting for up\&down and night scenarios, respectively. This highlights the crucial importance of LiDAR in 3D lane detection. In the up\&down case, our BEV representations exhibit a better capability in perceiving 3D lanes more accurately. Moreover, in the night case, where previous methods struggle due to imaging limitations, our multi-modal approach achieves remarkable performance. This is attributed to the effectiveness of our designed multi-modal framework, in which geometry-rich LiDAR can compensate for the deficiencies in low-quality visual information.  Similarly, when adopting a 0.5m distance threshold, we observe substantial performance improvements by equipping Persformer~\cite{chen2022persformer} with our LiDAR stream. This leads to an about 10-point increase in F-Score for both the 75m and 100m range settings, as illustrated in Table~\ref{tab:main_result_error}. These findings further reinforce the strength and effectiveness of the LiDAR stream.

\begin{table*}[ht]
    \centering
        \caption{Comprehensive comparison with other single/multi-modality methods on OpenLane \textbf{\texttt{val}} set. The evaluation is conducted using the original \textit{1.5m} threshold and valid range of LiDAR data, \textit{75m}. The errors on the X and Z axes are both measured in \textit{meters}.
        \textbf{Persformer$^\ast$:} we evaluate Persformer based on their provided checkpoint.
        \textbf{LiDAR$^{\dagger -}$:} our LiDAR stream uses only XYZ coordinates of points as inputs.
        \textbf{LiDAR$^{\dagger}$:} our LiDAR stream additionally uses point intensity and elongation as inputs.
        \textbf{$\ddagger$} denotes that we reimplement the models into the lane detection architecture based on their released codes.
        }
    \begin{tabular}{c|c!{\vrule width 1.2pt}c!{\vrule width 1.2pt}c|c|c|c}
    \toprule[1.2pt]
    Methods & Modality & F-Score$\uparrow$ & X error near$\downarrow$ & X error far$\downarrow$ & Z error near$\downarrow$ & Z error far$\downarrow$ \\
    \hline \hline
    Persformer~\cite{chen2022persformer}$^\ast$ & C & 53.3 & 0.478 & 0.563 & 0.363 & 0.408 \\
    \hline
    LiDAR$^{\dagger -}$  & L & 49.5 & 0.490 & 0.459 & 0.360 & 0.357 \\
    LiDAR$^\dagger$ & L & 53.8 & 0.462 & 0.435 & 0.332 & 0.343 \\
    \hline
    Persformer~\cite{chen2022persformer} + LiDAR$^\dagger$ & C+L & 56.9 & 0.453 & 0.467 & 0.354 & 0.357 \\
    PointPainting~\cite{vora2020pointpainting} RGB & C+L & 54.3 & 0.448 & 0.443 & 0.351 & 0.366 \\
    PointPainting~\cite{vora2020pointpainting} SEG& C+L & 51.5 & \textbf{0.376} & 0.392 & \textbf{0.303} & 0.330 \\
    BEVFusion~\cite{liang2022bevfusion}$^\ddagger$ & C+L & 54.7 & 0.455 & 0.431 & 0.319 & 0.334 \\
    SFD~\cite{wu2022sparse}$^\ddagger$ & C+L & 52.5 & 0.452 & 0.441 & 0.331 & 0.347 \\
    \hline
    M$^2$-3DLaneNet (ours) & C+L & \textbf{60.5} & 0.396 & \textbf{0.405} & 0.311  & \textbf{0.326} \\
    \toprule[1.2pt]
    \end{tabular}
    \vspace{-2mm}
    \label{tab:mm_result_comp}
\end{table*}
\label{subsec:mm-methods}
\subsubsection{Comparison with multi-modal methods}
In this section, we compare our method with other multi-modal techniques to further demonstrate the effectiveness of our proposed method. The comparison results are shown in Table~\ref{tab:mm_result_comp}, where the metrics are calculated under the 75m range setting. On top of Table~\ref{tab:mm_result_comp}, we introduce two baselines: LiDAR$^{\dagger-}$ and LiDAR$^{\dagger}$. These baselines utilize only the LiDAR branch from our~\modelname. For LiDAR$^{\dagger-}$, the inputs only contain only XYZ coordinates of points. In contrast, LiDAR$^{\dagger}$ additionally uses point intensity and elongation. As shown in~\Cref{tab:mm_result_comp}, LiDAR$^{\dagger-}$ does not exhibit an advantage over the pure image-based method. On the other hand, LiDAR$^{\dagger}$ shows a noticeable improvement over Persformer~\cite{chen2022persformer}, indicating that the {\textit{intensity is the main discriminative representation}} for lane detection on LiDARs. To gain a better understanding of how LiDAR contributes to 3D lane detection, we provide an intuitive visualization in~\Cref{fig:raw_lidar}. Except for the baselines,~\Cref{tab:mm_result_comp} also shows that equipping Persformer~\cite{chen2022persformer} with our LiDAR branch can boost its performance from 53.3 to 56.9 in F-Score, demonstrating the effectiveness of our LiDAR stream.

Furthermore, based on PointPainting~\cite{vora2020pointpainting}, we decorate the LiDAR point clouds with image features and feed the enhanced point clouds to our LiDAR branch for lane detection.
Besides,~\Cref{tab:mm_result_comp} shows two variants of the PointPainting models (PointPainting RGB and PointPainting SEG). The former directly decorates the point clouds with RGB colors from images, while the latter uses the pixel labels predicted by a pretrained 2D lane segmentation model~\cite{romera2017erfnet}. The poor performance of the SEG version may be attributed to inadequate segmentation results of the pretrained model. This issue arises due to the domain gap between OpenLane images and the training set of the segmentation models, as well as the inherent difficulty of lane segmentation. Finally, we compare our model with BEVFusion~\cite{liang2022bevfusion} and SFD~\cite{wu2022sparse}, both of which fuse the multi-modal features in a unified 3D space. While we are the first to explore multi-modal exclusively for 3D lane detection, these multi-modal methods cannot be directly applied to this task. To overcome this, we adapt these approaches to lane detection by replacing their task-specific heads with ours. While BEVFusion~\cite{liang2022bevfusion} and SFD~\cite{wu2022sparse} have shown promising results in 3D object detection, our investigations reveal that their performances are inferior to our approach when directly applied to the lane detection task. This disparity can be attributed to their lack of consideration for the unique characteristics of lanes, such as their slenderness and sparsity, which are crucial factors for achieving accurate lane detection. It is worth noting that SFD did not surpass LiDAR$^\dagger$ in terms of F-Score, potentially indicating its sensitivity to the accuracy of depth estimation.
\begin{table}[hb]\centering
\footnotesize
\setlength{\tabcolsep}{6pt}
\caption{\textbf{Ablation studies on designed modules.} All the ablation experiments are based on the multi-modal framework. \textbf{CS-DA}: Cross-Scale Deformation Attention. \textbf{CS-LDA}: Cross-Scale Line-restricted Deformable Attention module. \textbf{NBC}: the nearest BEV completion.}
\vspace{-2mm}
\resizebox{0.47\textwidth}{!}{
\begin{tabular}{ccc|c|c|c}
\toprule[1.2pt]
CS- & CS- & \multirow{2}{*}{NBC} & \multirow{2}{*}{F-Score} & {X error} & {Z error} \\
DA & LDA & & & near $\vert$ far & near $\vert$ far\\
\hline\hline
& & & 59.1 & 0.412 $\vert$ 0.420 & 0.321 $\vert$ 0.337 \\
& & \checkmark & 60.2 (+1.1) & 0.401 $\vert$ 0.403 & 0.312 $\vert$ 0.325 \\
& \checkmark &  & 60.3 (+1.2) & 0.403 $\vert$ 0.409 & 0.315 $\vert$ 0.332 \\
\checkmark & &  & 60.1 (+1.0) & 0.402 $\vert$ 0.413 & 0.319 $\vert$ 0.331 \\
 & \checkmark & \checkmark & \textbf{60.5} (+1.4) & 0.396 $\vert$ 0.405 & 0.311 $\vert$ 0.326 \\
\toprule[1.2pt]
\end{tabular}
}
\vspace{-6mm}
\label{tab:ablation_components}
\end{table}
\subsection{Design Analysis}
To understand the effectiveness of our proposed modules, we conduct ablation studies on each component in~\modelname, as summarized in Table~\ref{tab:ablation_components}.
After independently adopting the CS-LDA and NBC, the performance will be increased by 1.1 and 1.2, respectively. By employing both techniques simultaneously,~\modelname~achieves 60.5 (+1.4).
Notably, the joint adoption of both CS-LDA and NBC leads to a decrease in error and an increase in F-Score, indicating that their combination improves detection results and enhances localization accuracy.
Additionally, we present the results of the original deform attention, which causes a 0.2 drop in F-Score compared to CS-LDA.

\subsection{Model Complexity}
We summarize the parameters of models and test the FPS of them on one Nvidia Tesla V100-32G GPU, as shown in Table~\ref{tab:fps}.  We notice that our model uses fewer parameters compared to Persformer~\cite{chen2022persformer}, although we adopt two different modalities as inputs. In other words, we adopt a lightweight branch, about 7.5M parameters (the last line in Table~\ref{tab:fps}), to encode the 3D information. A stronger LiDAR backbone may further boost the performance. Note that 2D auxiliary heads are not included for both Persformer~\cite{chen2022persformer} and our~\modelname, as they are used to help the training. 
\begin{table}[ht]
\small
    \centering
    \caption{LiDAR Backbone$^\dagger$: the adopted LiDAR backbone in our~\modelname~model. For FPS, we run 1000 times forward and take the average value.}
    \begin{tabu}{c|c|c}
    \specialrule{0em}{3pt}{3pt}
    \hline\hline
        Model & \# Param & FPS \\
        \tabucline[1.2pt]{-}
        Persformer & 37.32M & 14.19 \\
        M$^2$-3DLaneNet & 29.71M & 8.75 \\
        \hline
        LiDAR Backbone$^\dagger$ & 7.46M & 47.34 \\
        \hline\hline
    \end{tabu}  
    \label{tab:fps}
\end{table}
\section{Conclusions}
In this work, we conduct comprehensive experiments to explore the effect of LiDAR on 3D lane detection and we present the~\modelname. 
It is a novel multi-modal 3D lane detection framework that utilizes both camera and LiDAR data simultaneously. 
By lifting 2D features through the generated dense depth map and fusing features in BEV space, \modelname~effectively extracts useful features from different modalities and integrates them for better 3D lane detection.
The architecture is validated on the OpenLane dataset and outperforms previous works.

\section{Acknowledgment}
This work was supported in part by NSFC-Youth 61902335,  by HZQB-KCZYZ-2021067, by the National Key R\&D Program of China with grant No.2018YFB1800800, by Shenzhen Outstanding Talents Training Fund, by Guangdong Research  Project No.2017ZT07X152, by Guangdong Regional Joint Fund-Key Projects 2019B1515120039,  by the  NSFC 61931024\&81922046, by helixon biotechnology company Fund and CCF-Tencent Open Fund.

{\small
\bibliographystyle{ieee_fullname}
\bibliography{egbib}
}

\end{document}